\documentclass[10pt,twocolumn,letterpaper]{article}

\usepackage{wacv}
\usepackage{times}
\usepackage{epsfig}
\usepackage{graphicx}
\usepackage{amsmath}
\usepackage{amssymb}

\usepackage[english]{babel}
\usepackage[utf8]{inputenc}
\usepackage{comment}
\usepackage{textcomp}
\usepackage[percent]{overpic}

\usepackage[skip=6pt,font=normalsize,labelsep=period]{caption}
\usepackage{tabularx}
\newcolumntype{Y}{>{\centering\arraybackslash}X}

\usepackage{color, colortbl}
\definecolor{black}{gray}{0.0}

\newcommand{\light}{\textcolor[rgb]{0.8,0.8,0.8}}
\newcommand{\tless}{T-LESS} 

\def\eg{\emph{e.g.}} 
\def\ie{\emph{i.e.}}

\usepackage[draft]{hyperref} 

\wacvfinalcopy 

\def\wacvPaperID{185} 

\ifwacvfinal\fi
\begin{document}

\title{\tless{}: An RGB-D Dataset for 6D Pose Estimation of Texture-less Objects}


\author{\hspace*{-1.0ex}Tomáš Hodaň$^{1}$, Pavel Haluza$^{1}$, Štěpán Obdržálek$^{1}$, Jiří Matas$^{1}$, Manolis Lourakis$^{2}$, Xenophon Zabulis$^{2}$ \vspace{0.5em}\\
\hspace*{-1.5ex}$^{1}$Center for Machine Perception, Czech Technical University in Prague, Czech Republic\\
\hspace*{-1.5ex}$^{2}$Institute of Computer Science, Foundation for Research and Technology -- Hellas, Heraklion, Greece\\
}

\maketitle
\ifwacvfinal\thispagestyle{empty}\fi

\begin{abstract}
\vspace{-0.5em}
We introduce \tless, a new public dataset for estimating the 6D pose,
i.e. translation and rotation, of texture-less rigid objects.
The dataset features thirty industry-relevant objects with no significant texture and no discriminative color or reflectance properties. The objects exhibit symmetries and mutual similarities in shape and/or size.
Compared to other datasets, a unique property is that some of the objects are parts of others.
The dataset includes training and test images that were captured with three synchronized sensors, specifically a structured-light and a time-of-flight RGB-D sensor and a high-resolution RGB camera. There are approximately 39K training and 10K test images from each sensor. Additionally, two types of 3D models are provided for each object, i.e. a manually created CAD model and a semi-automatically reconstructed one.
Training images depict individual objects against a black background. Test images originate from twenty test scenes having varying complexity,
which increases from simple scenes with several isolated objects to very challenging ones with multiple instances of several objects and with a high
amount of clutter and occlusion.
The images were captured from a systematically sampled view sphere around the object/scene,
and are annotated with accurate ground truth 6D poses of all modeled objects.
Initial evaluation results
indicate that the state of the art in 6D object pose estimation has ample room for improvement, especially in difficult cases with significant occlusion.
The \tless{} dataset is available online at \url{cmp.felk.cvut.cz/t-less}.
\end{abstract}

\section{Introduction}
\label{sec:introduction}

\begin{figure}[!t]
\begin{center}
    \begingroup
    \setlength{\tabcolsep}{1.5pt} 
    \renewcommand{\arraystretch}{0.9} 
    \begin{tabular}{ c c }
    \includegraphics[width=0.49\columnwidth]{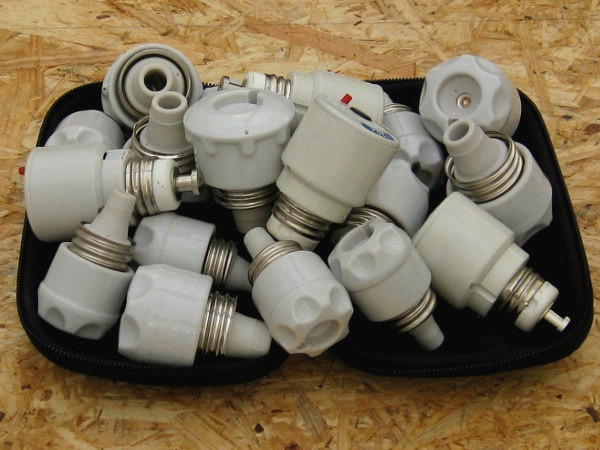} &
    \includegraphics[width=0.49\columnwidth]{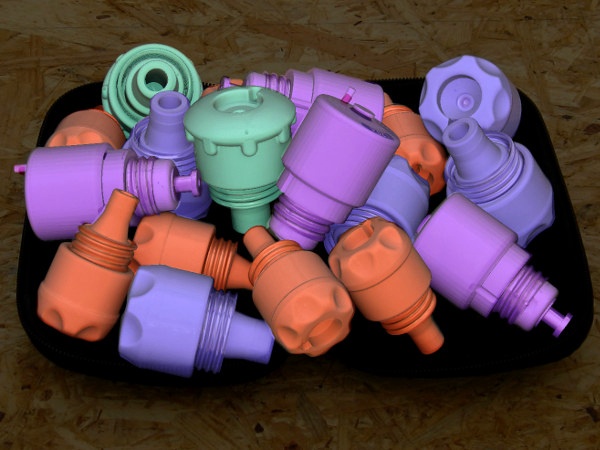} \\

    \includegraphics[width=0.49\columnwidth]{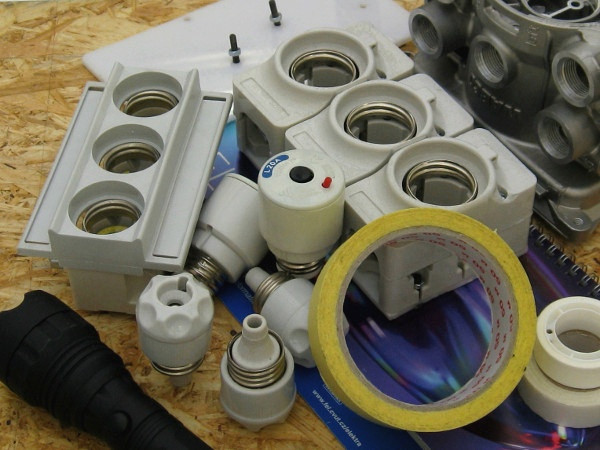} &
    \includegraphics[width=0.49\columnwidth]{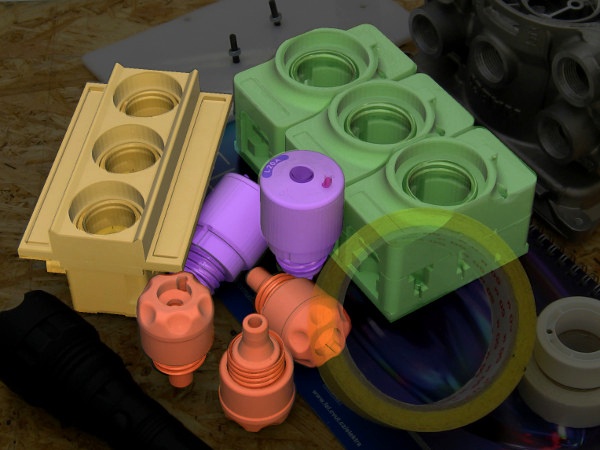} \\

    \includegraphics[width=0.49\columnwidth]{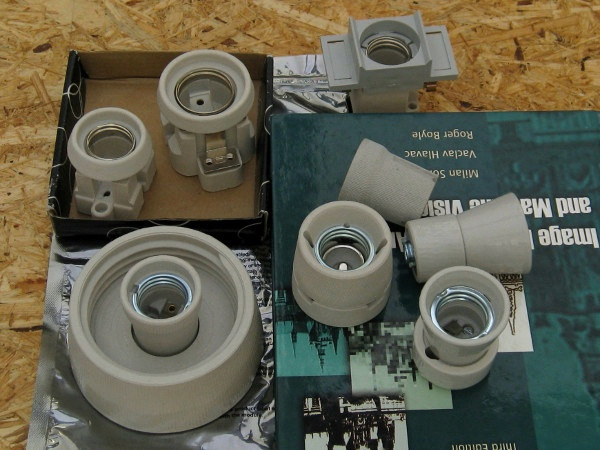} &
    \includegraphics[width=0.49\columnwidth]{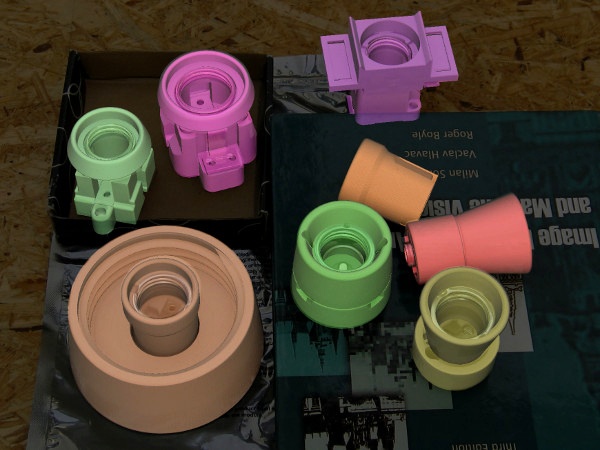}
    \end{tabular}
    \endgroup
    \caption{\label{fig:tless_intro} Examples of \tless~test images (left) overlaid with colored 3D object models at the ground truth 6D poses (right). Instances of the same object have the same color. The goal is to find instances of the modeled objects and estimate their 6D poses.}
\end{center}
\end{figure}


Texture-less rigid objects are common in human environments and the need to learn, detect and accurately localize them from images arises in a variety of applications. The pose of a rigid object has six degrees of freedom, \ie{} three in translation and three in rotation, and its full knowledge is often required.
In robotics, for example, the 6D object pose facilitates spatial reasoning and allows an end-effector to act upon an object. In an augmented reality scenario, object pose can be used to enhance one's perception of reality by
augmenting objects with extra information such as hints for assembly guidance.

The visual appearance of a texture-less object is dominated by its global shape, color, reflectance properties, and the configuration of light sources. The lack of texture implies that the object cannot be reliably recognized with traditional techniques relying on photometric local patch detectors and descriptors~\cite{collet2011moped,Lourakis13model}.
Instead, recent approaches that can deal with texture-less objects have focused on local 3D feature description~\cite{mian2010repeatability,tombari2010unique,guo2016comprehensive}, and semi-global or global description relying primarily on intensity edges and depth cues~\cite{hinterstoisser2012model,
Hodan2015detection,wohlhart2015learning,brachmann2016uncertainty,doumanoglou2016recovering,hinterstoisser2016going,kehl2016deep}.
Therefore, RGB-D data consisting of aligned color and depth images, obtained with
widely available Kinect-like sensors, have come to play an important role.

\begin{figure}[t!]
\begin{center}
\begingroup
\setlength{\tabcolsep}{0.0pt} 
\renewcommand{\arraystretch}{0.0} 

\begin{tabular}{ c c c c c c }
\rowcolor{black}
\begin{overpic}[width=0.1667\columnwidth]{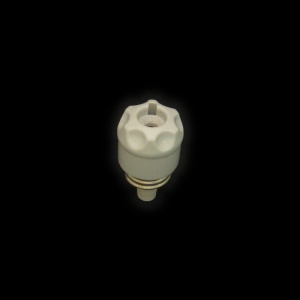}
\put(3,82){\scriptsize$\displaystyle\textbf{\light{1}}$}\end{overpic} &
\begin{overpic}[width=0.1667\columnwidth]{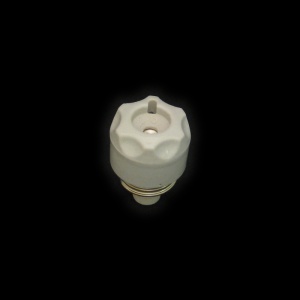}
\put(3,82){\scriptsize$\displaystyle\textbf{\light{2}}$}\end{overpic} &
\begin{overpic}[width=0.1667\columnwidth]{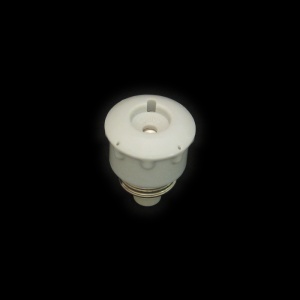}
\put(3,82){\scriptsize$\displaystyle\textbf{\light{3}}$}\end{overpic} &
\begin{overpic}[width=0.1667\columnwidth]{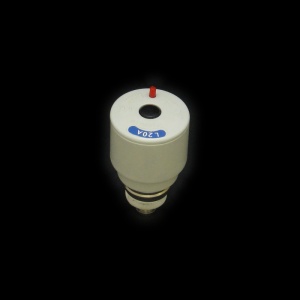}
\put(3,82){\scriptsize$\displaystyle\textbf{\light{4}}$}\end{overpic} &
\begin{overpic}[width=0.1667\columnwidth]{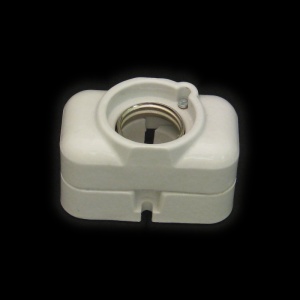}
\put(3,82){\scriptsize$\displaystyle\textbf{\light{5}}$}\end{overpic} &
\begin{overpic}[width=0.1667\columnwidth]{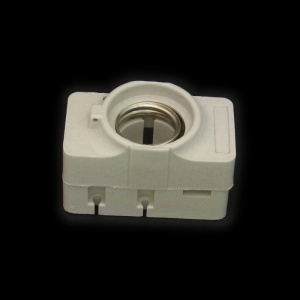}
\put(3,82){\scriptsize$\displaystyle\textbf{\light{6}}$}\end{overpic} \\

\rowcolor{black}
\begin{overpic}[width=0.1667\columnwidth]{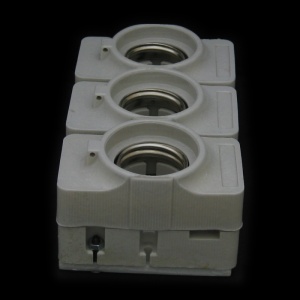}
\put(3,82){\scriptsize$\displaystyle\textbf{\light{7}}$}\end{overpic} &
\begin{overpic}[width=0.1667\columnwidth]{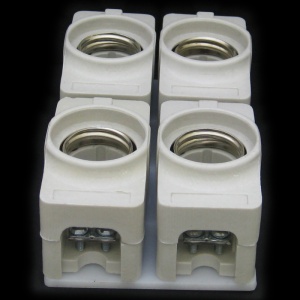}
\put(3,82){\scriptsize$\displaystyle\textbf{\light{8}}$}\end{overpic} &
\begin{overpic}[width=0.1667\columnwidth]{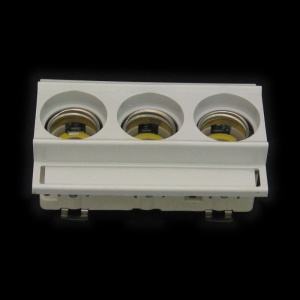}
\put(3,82){\scriptsize$\displaystyle\textbf{\light{9}}$}\end{overpic} &
\begin{overpic}[width=0.1667\columnwidth]{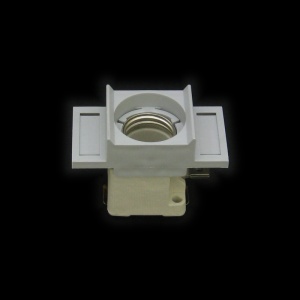}
\put(3,82){\scriptsize$\displaystyle\textbf{\light{10}}$}\end{overpic} &
\begin{overpic}[width=0.1667\columnwidth]{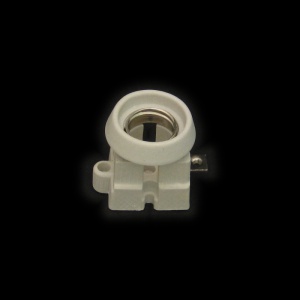}
\put(3,82){\scriptsize$\displaystyle\textbf{\light{11}}$}\end{overpic} &
\begin{overpic}[width=0.1667\columnwidth]{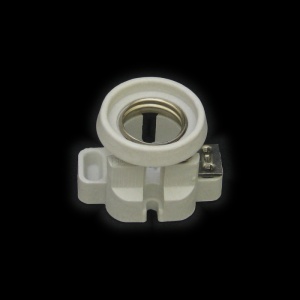}
\put(3,82){\scriptsize$\displaystyle\textbf{\light{12}}$}\end{overpic} \\

\rowcolor{black}
\begin{overpic}[width=0.1667\columnwidth]{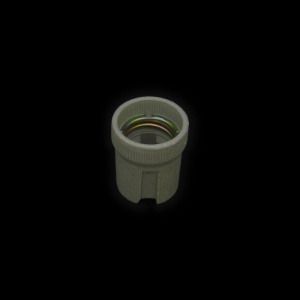}
\put(3,82){\scriptsize$\displaystyle\textbf{\light{13}}$}\end{overpic} &
\begin{overpic}[width=0.1667\columnwidth]{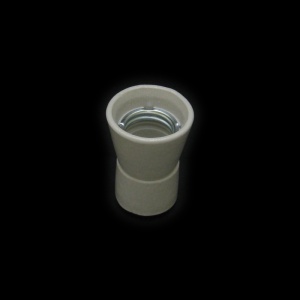}
\put(3,82){\scriptsize$\displaystyle\textbf{\light{14}}$}\end{overpic} &
\begin{overpic}[width=0.1667\columnwidth]{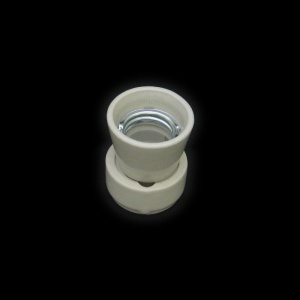}
\put(3,82){\scriptsize$\displaystyle\textbf{\light{15}}$}\end{overpic} &
\begin{overpic}[width=0.1667\columnwidth]{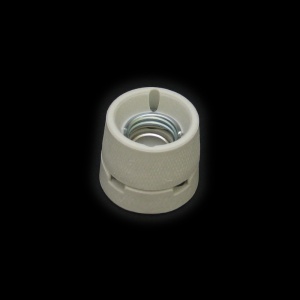}
\put(3,82){\scriptsize$\displaystyle\textbf{\light{16}}$}\end{overpic} &
\begin{overpic}[width=0.1667\columnwidth]{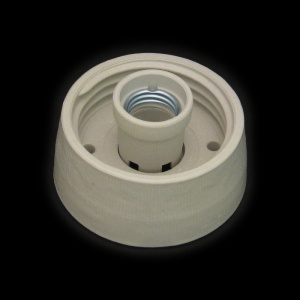}
\put(3,82){\scriptsize$\displaystyle\textbf{\light{17}}$}\end{overpic} &
\begin{overpic}[width=0.1667\columnwidth]{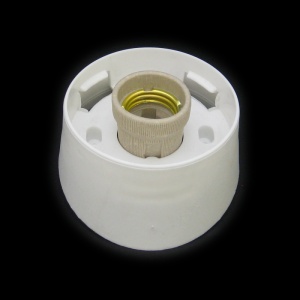}
\put(3,82){\scriptsize$\displaystyle\textbf{\light{18}}$}\end{overpic} \\

\rowcolor{black}
\begin{overpic}[width=0.1667\columnwidth]{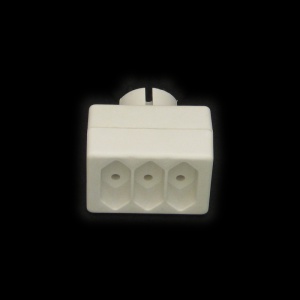}
\put(3,82){\scriptsize$\displaystyle\textbf{\light{19}}$}\end{overpic} &
\begin{overpic}[width=0.1667\columnwidth]{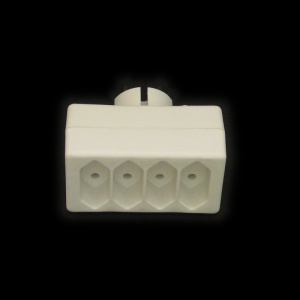}
\put(3,82){\scriptsize$\displaystyle\textbf{\light{20}}$}\end{overpic} &
\begin{overpic}[width=0.1667\columnwidth]{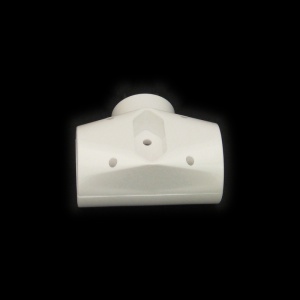}
\put(3,82){\scriptsize$\displaystyle\textbf{\light{21}}$}\end{overpic} &
\begin{overpic}[width=0.1667\columnwidth]{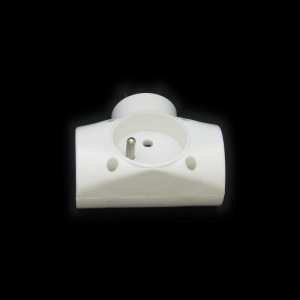}
\put(3,82){\scriptsize$\displaystyle\textbf{\light{22}}$}\end{overpic} &
\begin{overpic}[width=0.1667\columnwidth]{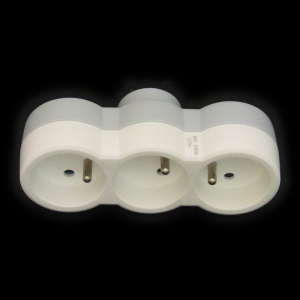}
\put(3,82){\scriptsize$\displaystyle\textbf{\light{23}}$}\end{overpic} &
\begin{overpic}[width=0.1667\columnwidth]{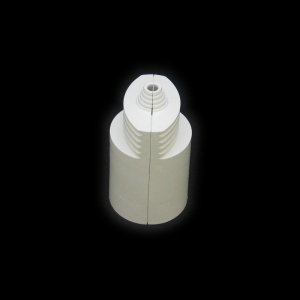}
\put(3,82){\scriptsize$\displaystyle\textbf{\light{24}}$}\end{overpic} \\

\rowcolor{black}
\begin{overpic}[width=0.1667\columnwidth]{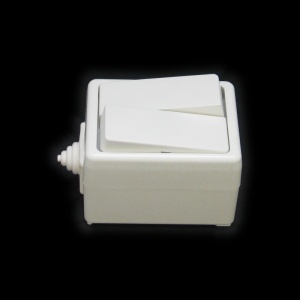}
\put(3,82){\scriptsize$\displaystyle\textbf{\light{25}}$}\end{overpic} &
\begin{overpic}[width=0.1667\columnwidth]{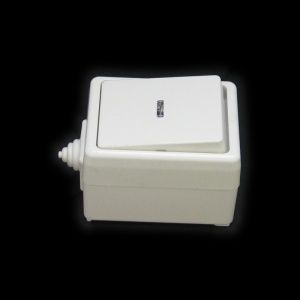}
\put(3,82){\scriptsize$\displaystyle\textbf{\light{26}}$}\end{overpic} &
\begin{overpic}[width=0.1667\columnwidth]{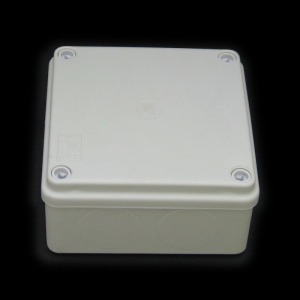}
\put(3,82){\scriptsize$\displaystyle\textbf{\light{27}}$}\end{overpic} &
\begin{overpic}[width=0.1667\columnwidth]{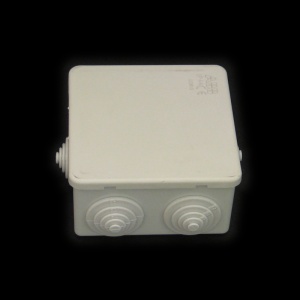}
\put(3,82){\scriptsize$\displaystyle\textbf{\light{28}}$}\end{overpic} &
\begin{overpic}[width=0.1667\columnwidth]{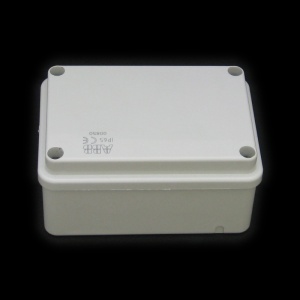}
\put(3,82){\scriptsize$\displaystyle\textbf{\light{29}}$}\end{overpic} &
\begin{overpic}[width=0.1667\columnwidth]{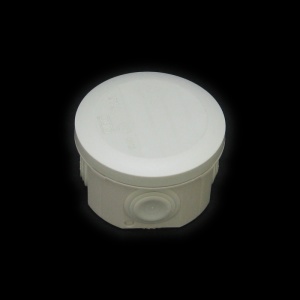}
\put(3,82){\scriptsize$\displaystyle\textbf{\light{30}}$}\end{overpic}

\end{tabular}

\vspace{1.5ex}


\begin{tabular}{ c c c c }

\rowcolor{black}
\begin{overpic}[width=0.25\columnwidth]{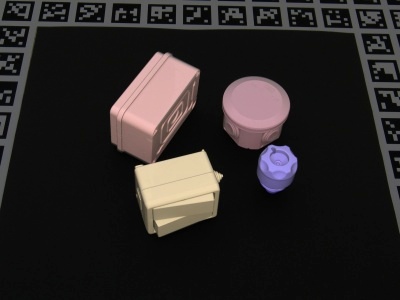}
\put(3,63){\scriptsize$\displaystyle\textbf{\light{1}}$}\end{overpic} &
\begin{overpic}[width=0.25\columnwidth]{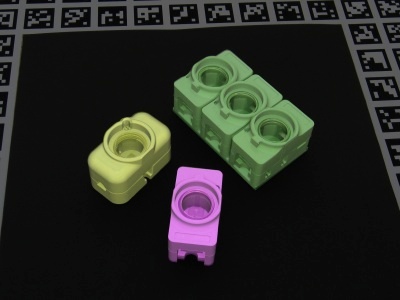}
\put(3,63){\scriptsize$\displaystyle\textbf{\light{2}}$}\end{overpic} &
\begin{overpic}[width=0.25\columnwidth]{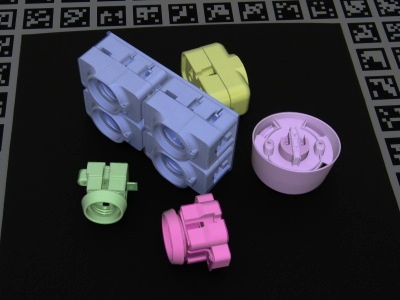}
\put(3,63){\scriptsize$\displaystyle\textbf{\light{3}}$}\end{overpic} &
\begin{overpic}[width=0.25\columnwidth]{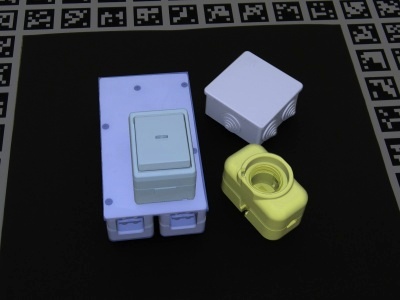}
\put(3,63){\scriptsize$\displaystyle\textbf{\light{4}}$}\end{overpic} \\

\rowcolor{black}
\begin{overpic}[width=0.25\columnwidth]{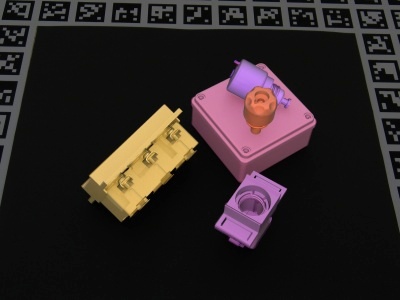}
\put(3,63){\scriptsize$\displaystyle\textbf{\light{5}}$}\end{overpic} &
\begin{overpic}[width=0.25\columnwidth]{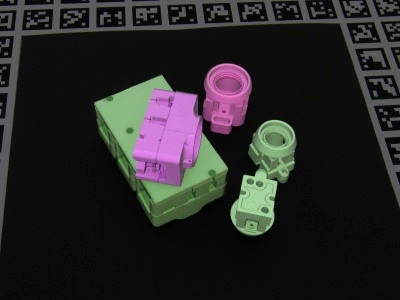}
\put(3,63){\scriptsize$\displaystyle\textbf{\light{6}}$}\end{overpic} &
\begin{overpic}[width=0.25\columnwidth]{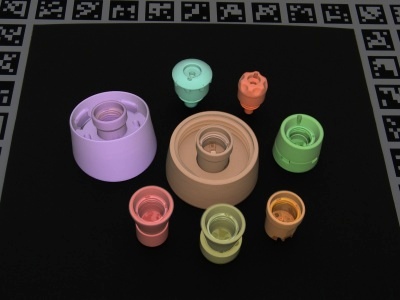}
\put(3,63){\scriptsize$\displaystyle\textbf{\light{7}}$}\end{overpic} &
\begin{overpic}[width=0.25\columnwidth]{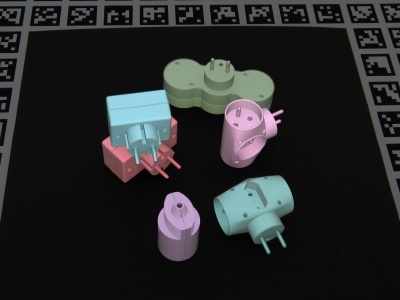}
\put(3,63){\scriptsize$\displaystyle\textbf{\light{8}}$}\end{overpic} \\

\rowcolor{black}
\begin{overpic}[width=0.25\columnwidth]{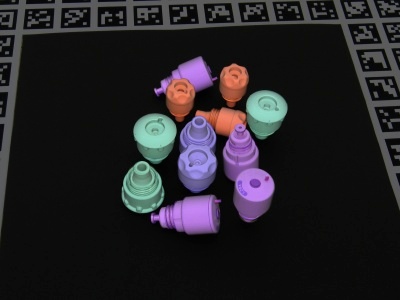}
\put(3,63){\scriptsize$\displaystyle\textbf{\light{9}}$}\end{overpic} &
\begin{overpic}[width=0.25\columnwidth]{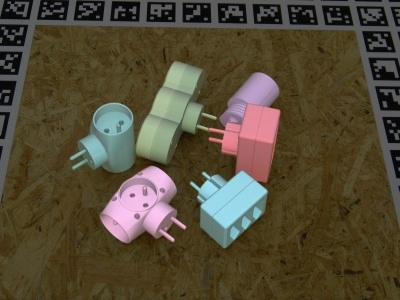}
\put(3,63){\scriptsize$\displaystyle\textbf{\light{10}}$}\end{overpic} &
\begin{overpic}[width=0.25\columnwidth]{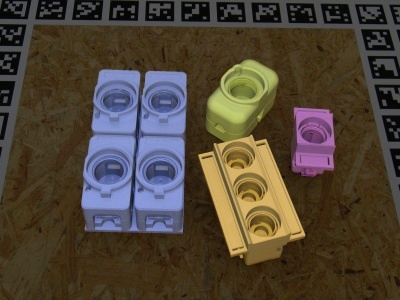}
\put(3,63){\scriptsize$\displaystyle\textbf{\light{11}}$}\end{overpic} &
\begin{overpic}[width=0.25\columnwidth]{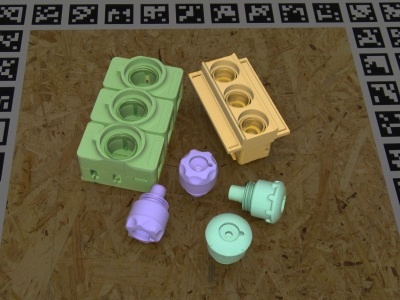}
\put(3,63){\scriptsize$\displaystyle\textbf{\light{12}}$}\end{overpic} \\

\rowcolor{black}
\begin{overpic}[width=0.25\columnwidth]{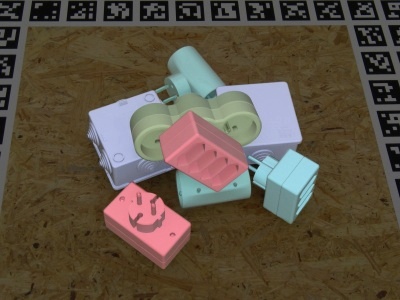}
\put(3,63){\scriptsize$\displaystyle\textbf{\light{13}}$}\end{overpic} &
\begin{overpic}[width=0.25\columnwidth]{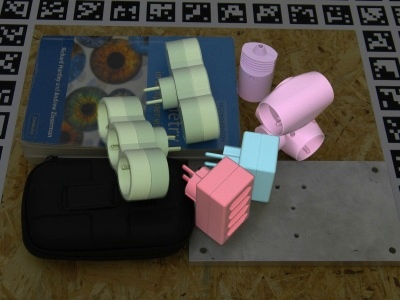}
\put(3,63){\scriptsize$\displaystyle\textbf{\light{14}}$}\end{overpic} &
\begin{overpic}[width=0.25\columnwidth]{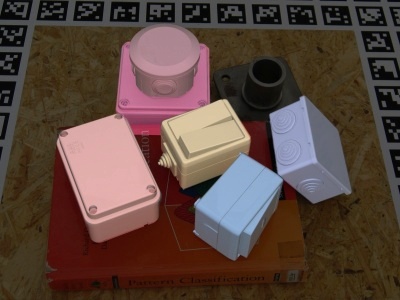}
\put(3,63){\scriptsize$\displaystyle\textbf{\light{15}}$}\end{overpic} &
\begin{overpic}[width=0.25\columnwidth]{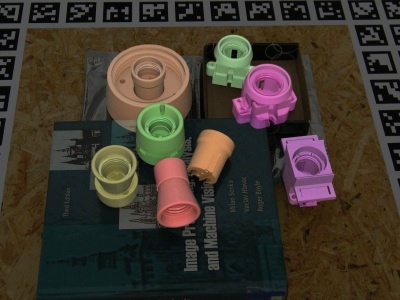}
\put(3,63){\scriptsize$\displaystyle\textbf{\light{16}}$}\end{overpic} \\

\rowcolor{black}
\begin{overpic}[width=0.25\columnwidth]{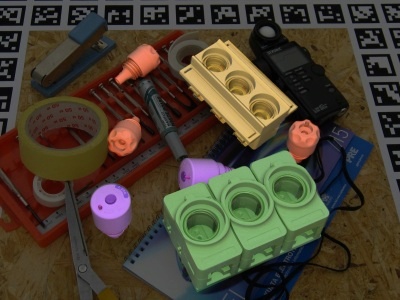}
\put(3,63){\scriptsize$\displaystyle\textbf{\light{17}}$}\end{overpic} &
\begin{overpic}[width=0.25\columnwidth]{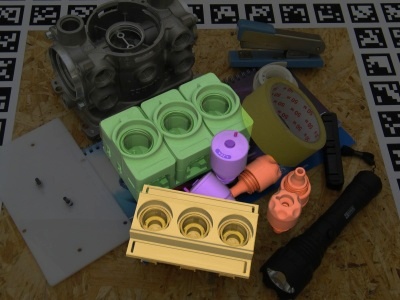}
\put(3,63){\scriptsize$\displaystyle\textbf{\light{18}}$}\end{overpic} &
\begin{overpic}[width=0.25\columnwidth]{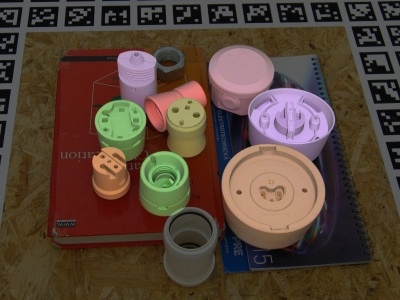}
\put(3,63){\scriptsize$\displaystyle\textbf{\light{19}}$}\end{overpic} &
\begin{overpic}[width=0.25\columnwidth]{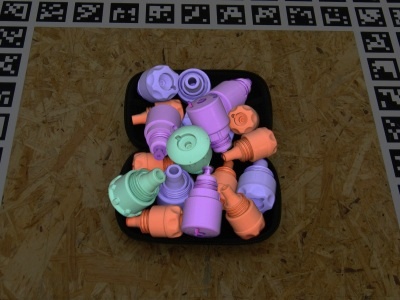}
\put(3,63){\scriptsize$\displaystyle\textbf{\light{20}}$}\end{overpic}
\end{tabular}
\endgroup
\vspace{2.0pt}
\caption{\label{fig:tless_overview} \tless{} includes training images and 3D models of 30 objects (top) and test images of 20 scenes (bottom) -- shown overlaid with colored 3D object models at the ground truth poses. The images were captured from a systematically sampled view sphere around an object/scene and are annotated with accurate ground truth 6D poses of all modeled objects.}
\end{center}
\end{figure}

In this paper, we introduce a new public dataset for 6D pose estimation of texture-less rigid objects. An overview of the included objects and test scenes is provided in Fig.~\ref{fig:tless_overview}. The dataset features thirty commodity electrical parts which have no significant texture, discriminative color or distinctive reflectance properties, and often bear similarities in shape and/or size. Furthermore, a unique characteristic of the objects is that some of them are parts of others. For example, objects 7 and 8 are built up from object 6, object 9 is made of three copies of object 10 stacked next to each other, whilst the center part of objects 17 and 18 is nearly identical to object 13. Objects exhibiting similar properties are common in industrial environments. 


The dataset includes training and test images captured with a triplet of sensors, \ie{} a~structured light RGB-D sensor Primesense Carmine 1.09, 
a time-of-flight RGB-D sensor Microsoft Kinect v2, and an RGB camera Canon IXUS 950 IS. The sensors were time-synchronized and had similar perspectives. All images were obtained with an automatic procedure that systematically sampled images from a view sphere, resulting in \texttildelow39K training and \texttildelow10K test images from each sensor. The training images depict objects in isolation with a black background, while the test images originate from twenty table-top scenes with arbitrarily arranged objects. Complexity of the test scenes varies from those with several isolated objects and a clean background to very challenging ones with multiple instances of several objects and with a high amount of occlusion and clutter. Additionally, the dataset contains two types of 3D mesh models for each object; one manually created in CAD software and one semi-automatically reconstructed from the training RGB-D images. All occurrences of the modeled objects in the training and test images are annotated with accurate ground truth 6D poses; see Fig.~\ref{fig:tless_intro} for their qualitative and Sec.~\ref{sec:gtp_accur} for their quantitative evaluation.


The dataset is intended for evaluating various flavors of the 6D object pose estimation problem~\cite{Hodan2016evaluation} and other related problems, such as 2D object detection~\cite{tombari2013bold,Hodan2015efficient} and object segmentation~\cite{tombari2011online,georgakis2016multiview}. Since images from three sensors are available, one may also study the importance of different input modalities for a given problem. Another option is to use the training images for evaluating 3D object reconstruction methods~\cite{fastfusion2014}, where the provided CAD models can serve as the ground truth.

Our objectives in designing \tless{} were to provide a dataset of a substantial but manageable size, with a rigorous and complete ground truth annotation that is accurate to the level of sensor resolution, and with a significant variability in complexity, so that it would provide different levels of difficulty and be reasonably future-proof, \ie{} solvable, but not solved by the current state-of-the-art methods.
The difficulty of the dataset for 6D object pose estimation is demonstrated by the relatively low performance of the method by Hoda\v{n} et al.~\cite{Hodan2015detection}.
This method
otherwise achieves a performance close to the state of the art on the well-established dataset of Hinterstoisser et al.~\cite{hinterstoisser2012model}.

The remainder of the paper is organized as follows. Sec.~\ref{sec:related} reviews related datasets, Sec.~\ref{sec:dataset} describes technical details of
the acquisition and post-processing of the \tless{} dataset, Sec.~\ref{sec:expers} assesses the accuracy of the ground truth poses
and provides initial evaluation results, and Sec.~\ref{sec:concl} concludes the paper.

\section{Related Datasets} \label{sec:related}


First we review datasets for estimating the 6D pose of specific rigid objects, grouped by the type of provided images, then we mention a few datasets designed for similar problems. If not stated otherwise, these datasets supply ground truth annotations in the form of 6D object poses.

\subsection{RGB-D Datasets}

Only a few public RGB-D datasets, from over one hundred reported by Firman in~\cite{firman2016rgbd},
enable the evaluation of 6D object pose estimation methods.
Most of the datasets reviewed in this section were captured with Microsoft \mbox{Kinect v1} or Primesense \mbox{Carmine 1.09}, which represent the first generation of consumer-grade RGB-D sensors operating on the structured-light principle. The dataset introduced in~\cite{georgakis2016multiview} was captured with Microsoft \mbox{Kinect v2}, which is based on the time-of-flight principle.

For texture-less objects, the dataset of Hinterstoisser et al.~\cite{hinterstoisser2012model} has become a standard
benchmark used in most of the recent work, \eg{} \cite{rios2013discriminatively,brachman2014learning,tejani14latent,Hodan2015detection,wohlhart2015learning}. It contains 15 texture-less objects represented by a color 3D mesh model. Each object is associated with a test sequence consisting of \texttildelow1200 RGB-D images,
each of which includes exactly one instance of the object. 
The test sequences feature significant 2D and 3D clutter, but only mild occlusion, and since the objects have discriminative color, shape and/or size, their recognition is relatively easy. In the 6D localization problem (where information about the number and identity of objects present in the images is provided beforehand~\cite{Hodan2016evaluation}), state-of-the-art methods achieve recognition rates that exceed 95\% for most of the objects. Brachmann et al.~\cite{brachman2014learning} provided additional ground truth poses for all modeled objects in one of the test sequences from~\cite{hinterstoisser2012model}. This extended annotation introduces challenging test cases with various levels of occlusion and allows the evaluation of multiple object localization, with each object appearing in a single instance. 

Tejani et al.~\cite{tejani14latent} presented a dataset with 2 texture-less and 4 textured objects. For each object, a color 3D mesh model is provided together with a test sequence of over 700 RGB-D images. The images
show several object instances with no to moderate occlusion, and with 2D and 3D clutter. 
Doumanoglou et al.~\cite{doumanoglou2016recovering} provide a dataset with 183 test images of 2 textured objects from~\cite{tejani14latent} that appear in multiple instances in a challenging bin-picking scenario with heavy occlusion. Furthermore, they provide color 3D mesh models of another 6 textured objects and 170 test images depicting the objects placed on a kitchen table.


The Challenge and Willow datasets~\cite{xie2013multimodal}, which were collected for the 2011 ICRA Solutions in Perception Challenge, share a set of 35 textured household objects. Training data for each object is given in the form of 37 RGB-D training images
that show the object from different views, plus a color point cloud obtained by merging the training images. The Challenge and Willow datasets respectively contain 176 and 353 test RGB-D images of several objects in single instances
placed on top of a turntable. The Willow dataset also features distractor objects and object occlusion.
Similar is the TUW dataset~\cite{aldoma2014automation} that presents 17 textured and texture-less objects appearing in 224 test RGB-D images. 
Instead of a turntable setup, images were obtained by moving a robot around a static cluttered environment with some objects appearing in multiple instances.

The Rutgers dataset~\cite{Rennie2016} is focused on
perception for robotic manipulation during pick-and-place tasks and comprises of images
from a cluttered warehouse environment. It includes color 3D mesh models for 24 mostly textured objects from the Amazon Picking Challenge 2015~\cite{correll2016lessons}, that were captured in more than 10K test RGB-D images with various amounts of occlusion. 

Aldoma et al.~\cite{aldoma2012global} provide 3D mesh models without color information of 35 household objects that are both textured and texture-less and are often symmetric and mutually similar in shape and size. There are 50 test RGB-D images of
table-top scenes with multiple objects in single instances, with no clutter and various levels of occlusion.

The BigBIRD dataset~\cite{Singh2014} includes images of 125 mostly textured objects that were captured in isolation on a turntable with multiple calibrated
RGB-D and DSLR sensors. For each object, the
dataset provides 600 RGB-D point clouds, 600 high-resolution RGB images, and a color 3D mesh model reconstructed from the point clouds.
Since BigBIRD was acquired under very controlled conditions, it is not concerned with occlusions, clutter, lighting changes or varying object-sensor distance.
Georgakis et al.~\cite{georgakis2016multiview} provide
6735 test RGB-D images from kitchen scenes including a subset of the BigBIRD objects.
Ground truth for objects in the test images is provided only in the form of 2D bounding boxes and 3D point labeling.

Lai et al.~\cite{lai2011large} created an extensive dataset with 300 common household objects captured
on a turntable from three elevations.
It contains 250K segmented RGB-D images
and 22 annotated video sequences with a few hundred RGB-D frames in each.
Ground truth is provided only in the form of approximate rotation angles for training images and in the form of 3D point labeling for test images.

Schlette et al.~\cite{schlette2014new}
synthesized RGB-D images from simulated object manipulation scenarios involving 4 texture-less objects from the Cranfield assembly benchmark~\cite{collins1985development}. Several small datasets that were used for evaluation of the SHOT descriptor are provided by Salti et al.~\cite{Salti2014}. These datasets include synthetic data as well as data acquired with a spacetime-stereo method and an RGB-D sensor. 






\subsection{Depth-only and RGB-only Datasets}



The depth-only dataset of Mian et al.~\cite{mian2006three} includes 3D mesh models of 5 objects and 50 test depth images acquired with an industrial range scanner.
The test scenes contain only the modeled objects that occlude each other. A similar dataset is provided by Taati et al.~\cite{taati2007variable}. 
The Desk3D dataset~\cite{Bonde2014} comprises of 3D mesh models for 6 objects which are captured in over 850 test depth images with occlusion, clutter and similarly looking distractor objects. The dataset was obtained with 
an RGB-D sensor, however only the depth images
are publicly available.

The IKEA dataset by Lim et al.~\cite{lim2013parsing} provides RGB images with objects being aligned with their exactly matched 3D models. Crivellaro et al.~\cite{Crivellaro2015}
supply 3D CAD models and annotated RGB sequences 
with 3 highly occluded and texture-less objects. Mu\~{n}oz et al.~\cite{Munoz2016} provide RGB sequences of 6 texture-less objects that are each imaged in isolation against a clean background and without occlusion. 
Further to the above, there exist RGB datasets 
such as \cite{damen2012real,tombari2013bold,rios2013discriminatively,hsiao2014occlusion}, for which the ground truth is provided only in the form of 2D bounding boxes.

\subsection{Datasets for Similar Problems}

The RGB-D dataset of Michel et al.~\cite{michel2015pose} is focused on articulated objects, where the goal is to estimate the 6D pose of each object part, subject to the constraints introduced by their joints.
There are also datasets for categorical pose estimation.
For example, the 3DNet~\cite{wohlkinger20123dnet} and the UoB-HOOC~\cite{UoBHOOC2016} contain generic 3D models and RGB-D images annotated with 6D object poses. The UBC VRS~\cite{meger2011mobile}, the RMRC (a subset of NYU Depth v2~\cite{silberman2012indoor} with annotations derived from~\cite{guo2013support}), the B3DO~\cite{janoch2013category}, and the SUN RGB-D~\cite{song2015sun} provide no 3D models and ground truth only in the form of bounding boxes.
The PASCAL3D+~\cite{xiang2014beyond} and the ObjectNet3D~\cite{xiang2016objectnet3d} provide generic 3D models and ground truth 6D poses, but only RGB images.
\begin{figure}[t!]
	\begin{center}
		\begin{overpic}[width=0.75\columnwidth]{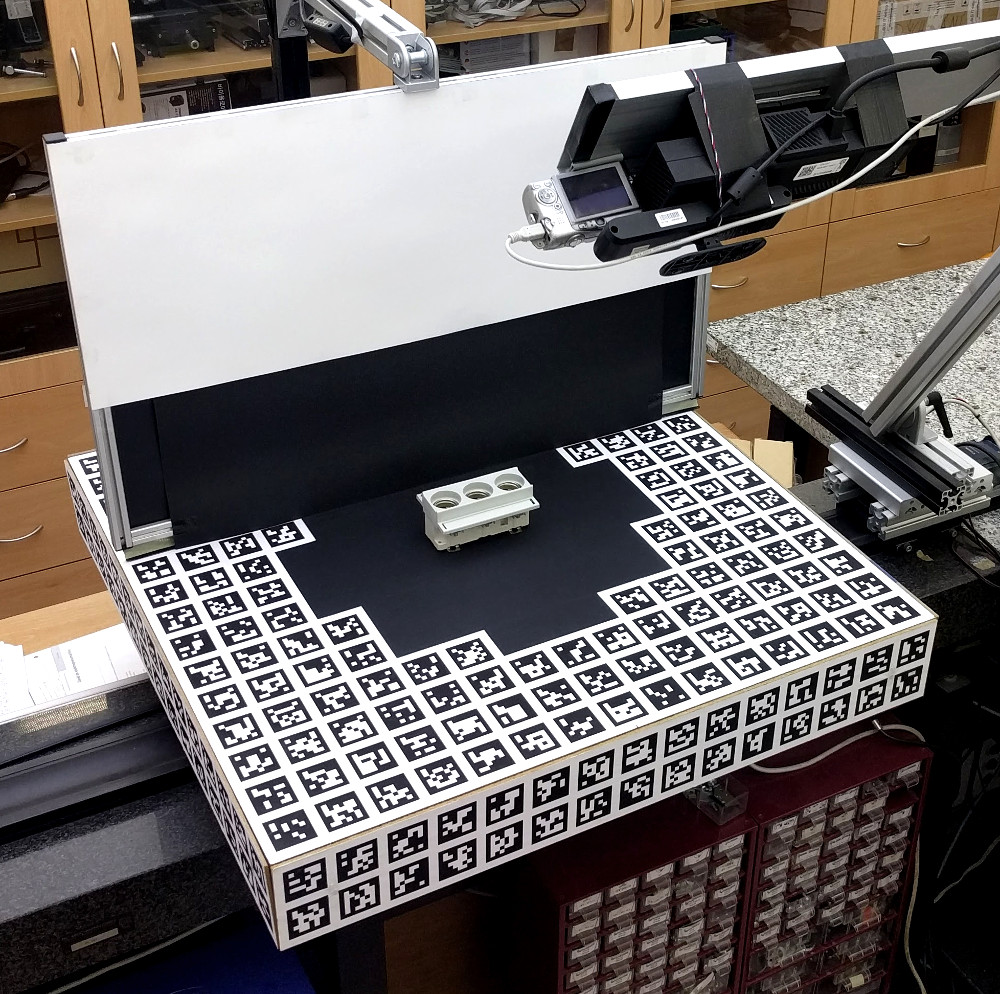}
			\put(80.5,81){\color{white}\circle*{10}}
			\put(78.9,78.7){\large$\displaystyle\textbf{3}$}
			\put(33,25){\color{white}\circle*{10}}
			\put(31.4,22.7){\large$\displaystyle\textbf{1}$}
			\put(34,55){\color{white}\circle*{10}}
			\put(32.4,52.7){\large$\displaystyle\textbf{2}$}
		\end{overpic}
		\vspace{2.0pt}
		\caption{\label{fig:capturing_rig} Acquisition setup: 1)~turntable with marker field, 2)~screen ensuring a black background for training images, removed when capturing test images, 3)~triplet of
sensors attached to a jig with adjustable tilt.}
	\end{center}
\end{figure}
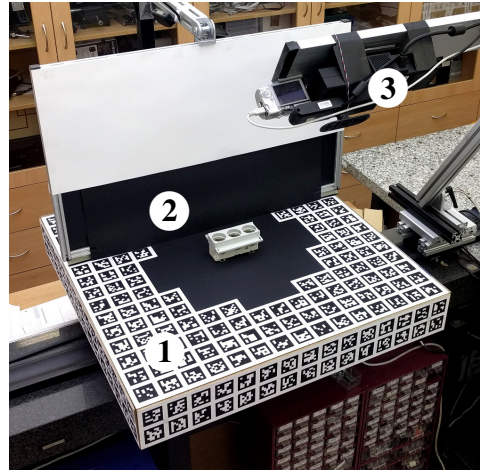
	
\begin{figure}[h!]
	\begin{center}
		\begingroup
		\setlength{\tabcolsep}{1.5pt} 
		\renewcommand{\arraystretch}{0.9} 
		\begin{tabular}{ c c c }
			\includegraphics[width=0.32\columnwidth]{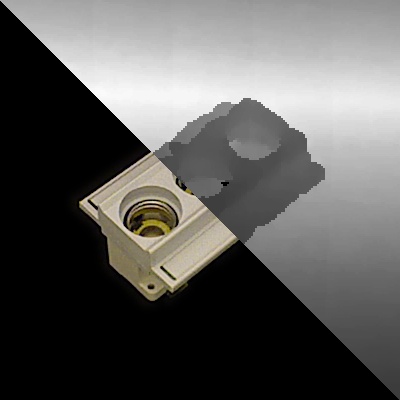} &
			\includegraphics[width=0.32\columnwidth]{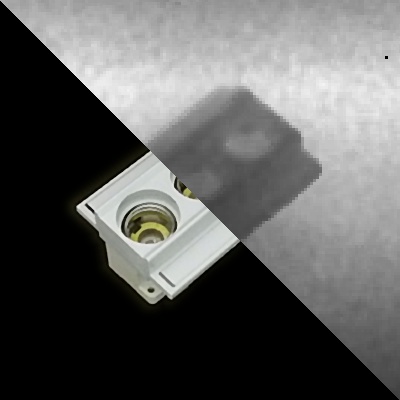} &
			\includegraphics[width=0.32\columnwidth]{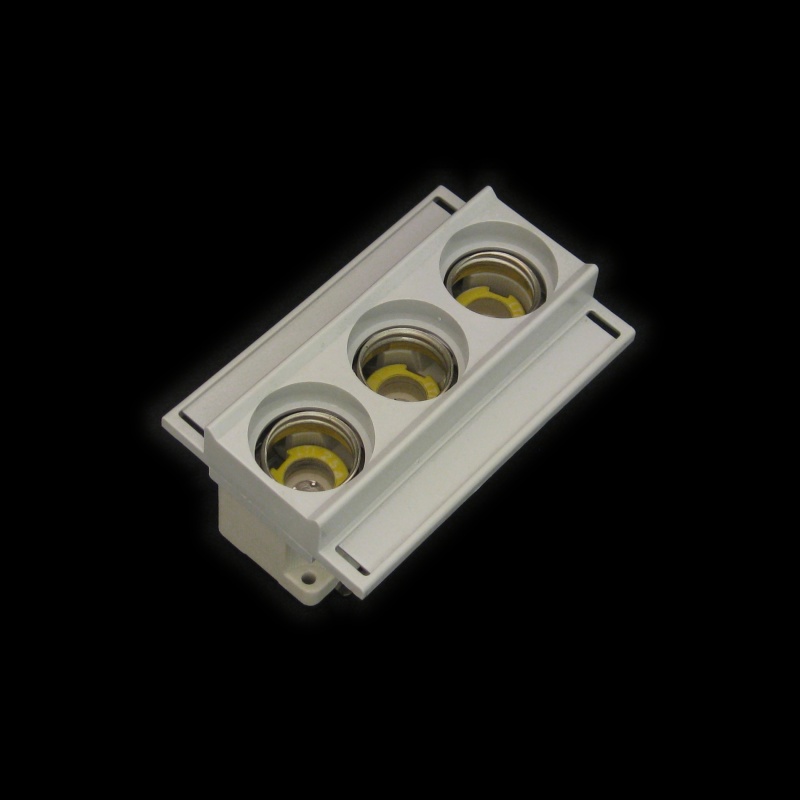} \\
			
			\includegraphics[width=0.32\columnwidth]{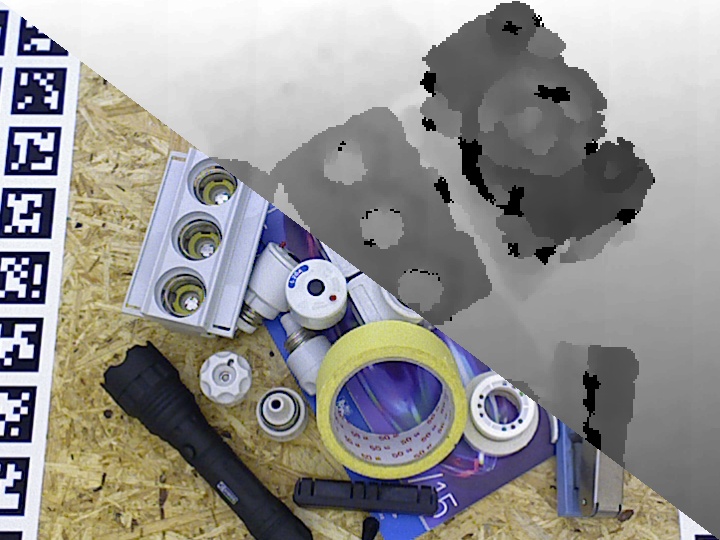} &
			\includegraphics[width=0.32\columnwidth]{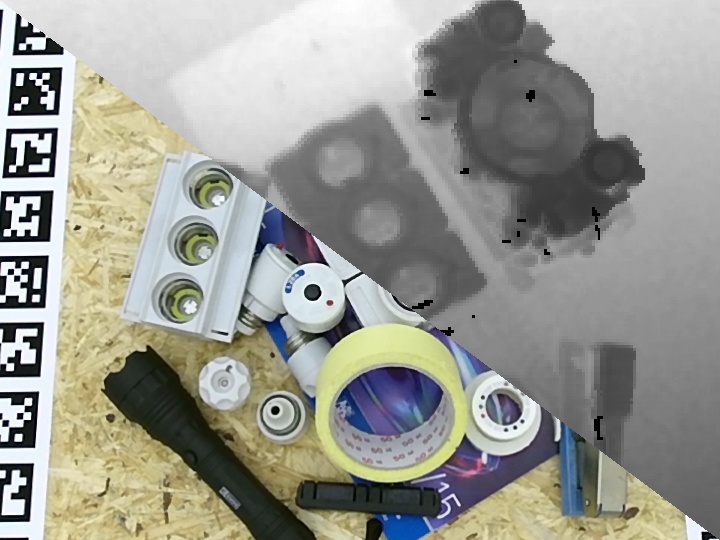} &
			\includegraphics[width=0.32\columnwidth]{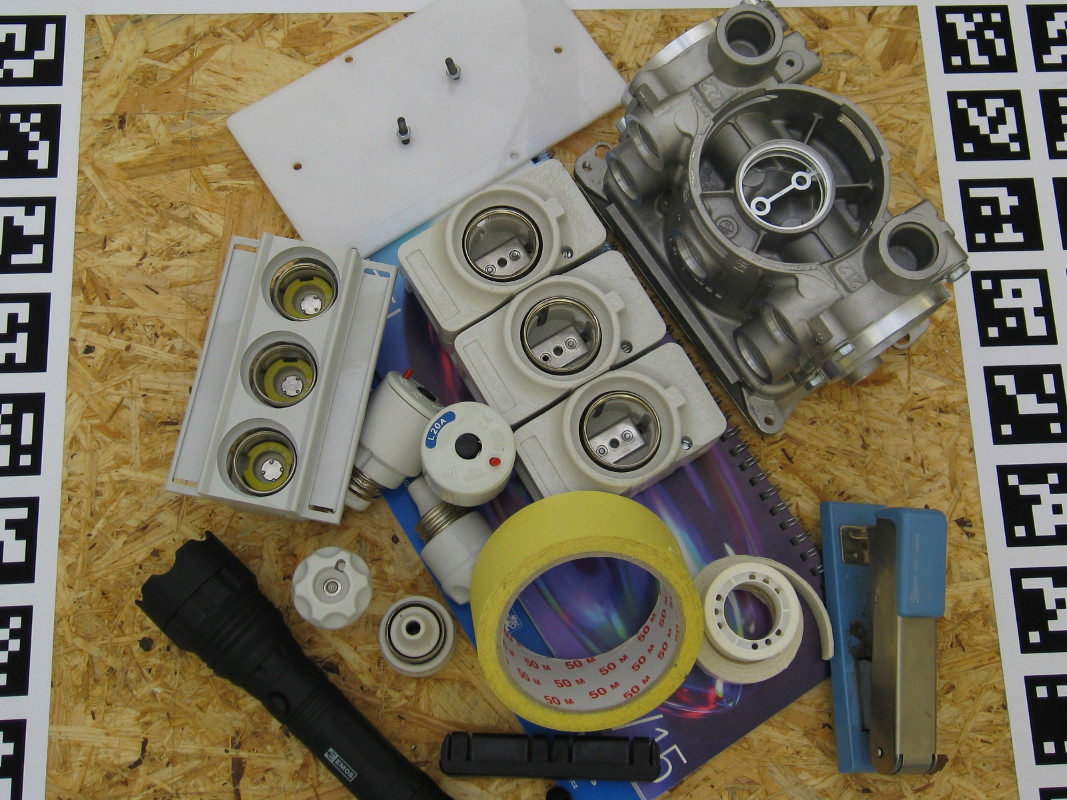}
		\end{tabular}
		\endgroup
		\caption{\label{fig:rgbd_views} Sample training (top) and test (bottom) images. Left: RGB-D images from Primesense Carmine 1.09. Middle: RGB-D images from Microsoft Kinect v2. Right: High-resolution RGB images from Canon IXUS 950 IS. For the RGB-D images, bottom-left halves show the RGB components whereas the top-right show the depth components.}
	\end{center}
\end{figure}

\section{The \tless{} Dataset}
\label{sec:dataset}


Compared to the reviewed datasets, \tless{} is unique in its combination of the following characteristics.
It contains 1) a larger number of industry-relevant objects, 2) training images captured under controlled conditions, 3) test images with large viewpoint changes, objects in multiple instances, affected by clutter and occlusion; including test cases that are challenging even for the state-of-the-art methods, 4) images captured with a synchronized and calibrated triplet of sensors, 5) accurate ground truth 6D poses for all modeled objects, and 6) two types of 3D models for each object.


The rest of the section describes the process of dataset preparation, which includes image acquisition, camera calibration, depth correction, 3D object model generation and the ground truth pose annotation.

\subsection{Acquisition Setup}
\label{sec:acquisition_setup}

The training and test images were captured with the aid of the setup shown in Fig.~\ref{fig:capturing_rig}. It consists of a turntable, where the imaged objects were placed, and a jig with adjustable tilt, to which the sensors were attached. A marker field used for camera pose estimation was affixed to the turntable. The field was extended vertically to the sides of the turntable to facilitate pose estimation at lower elevations.
To capture training images, the objects were placed in the middle of the turntable and in the front of a black screen, which ensured a uniform background at all elevations. To introduce a non-uniform background in the test images, a sheet of plywood with markers at its edges was placed on the top of the turntable. In some scenes, the objects were placed on the top of other objects (\eg~books) to give them different elevations and thus invalidate a ground plane assumption that might be made by an evaluated method.
The depth of object surfaces in the training and test images is in the range $0.53-0.92\,m$, which is within the sensing ranges of the used RGB-D sensors that are $0.35-1.4\,m$ for Carmine and $0.5-4.5\,m$ for Kinect.

\subsection{Calibration of Sensors}
\label{sec:calibration}

Intrinsic and distortion parameters of the sensors were estimated with the standard checkerboard-based procedure using OpenCV~\cite{bradski2008learning}.
The root mean square re-projection error calculated at corners of the calibration checkerboard squares is $0.51\,\text{px}$ for Carmine, $0.35\,\text{px}$ for Kinect, and $0.43\,\text{px}$ for Canon. For the RGB-D sensors, the calibration was performed with the RGB images. The depth images were aligned to the RGB images using the factory depth-to-color registration available through manufacturer's SDKs (OpenNI 2.2 and Kinect for Windows SDK 2.0). The color and aligned depth images, which are included in the dataset, are already processed to remove radial distortion. The intrinsic parameters can be found at the dataset website.

All sensors were synchronized and extrinsically calibrated with respect to the turntable, making it possible to register any pair of images. Synchronization was essential since the images were taken while the turntable was spinning. 
The extrinsic calibration was achieved using fiducial BCH-code markers from ARToolKitPlus~\cite{wagner2007artoolkitplus}. Specifically, the detection of 
particular markers in an image combined with the knowledge of their physical location on the turntable provided a set of 2D-3D correspondences. These were used to estimate the camera pose in the turntable coordinate system by robustly solving the PnP problem and then refining the estimated 6D pose by non-linearly minimizing the cumulative re-projection error with the posest library from~\cite{Lourakis13model}. 
The root mean square re-projection error, which was calculated at marker corners in all test images,
is $1.27\,\text{px}$ for Carmine, $1.37\,\text{px}$ for Kinect, and $1.50\,\text{px}$ for Canon. 
This measure combines errors in sensor calibration, marker field detection and sensor pose estimation and is therefore larger than the aforementioned error in sensor intrinsic calibration.


\subsection{Training and Test Images}
\label{sec:train_test_images}

A common strategy for dealing with poorly textured objects is to adopt a template-based approach trained on object images that are acquired with a dense sampling of viewpoints, \eg~\cite{damen2012real,hinterstoisser2012model,rios2013discriminatively,Hodan2015detection}. To support such approaches, \tless{} offers training images of every object in isolation from a full view sphere. These images were obtained with a systematic acquisition procedure which uniformly sampled elevation from $85^{\circ}$ to $-85^{\circ}$ with a $10^{\circ}$ step and the complete azimuth range with a $5^{\circ}$ step.
Views from the upper and lower hemispheres were captured separately, turning the object upside down in between. In total, there are
$18 \times 72 = 1296$
training images per object from each sensor.
Exceptions are objects 19 and 20, for which only views from the upper hemisphere were captured, specifically $648$ images from elevation $85^{\circ}$ to $5^{\circ}$. These objects are horizontally symmetric at the pose in which they were placed on the turntable, thus the views from the upper hemisphere are sufficient to capture their appearance.
Test scenes were captured from a view hemisphere with a $10^{\circ}$ step in elevation (ranging from $75^{\circ}$ to $15^{\circ}$) and a $5^{\circ}$ step in azimuth. A total of $7 \times 72 = 504$ test images were captured per scene by each sensor.

To remove irrelevant parts of the scene in the images periphery, the provided images are cropped versions of the captured ones. Resolution of the provided images is as follows: $400 \times 400\,\text{px}$ for training RGB-D images from Carmine and Kinect, $1900 \times 1900\,\text{px}$ for training RGB images from Canon, $720 \times 540\,\text{px}$ for test RGB-D images from Carmine and Kinect and $2560 \times 1920\,\text{px}$ for test RGB images from Canon. Sample images are shown in Fig.~\ref{fig:rgbd_views}.

Parts of the marker field were visible in some of the training images, especially at lower elevations. These were masked to ensure a black background everywhere around the objects. To achieve this, we identified an object mask in an image by back-projecting its CAD model and gradually darkened the image moving from the mask perimeter towards the image border. 

\subsection{Depth Correction}
\label{sec:depth_correction}

Similarly to~\cite{foix2011lock,sturm2012benchmark}, we observed that the depths measured by the RGB-D sensors exhibit a systematic error. To remove it, we collected depth measurements $d$ at projections of the marker corners and computed their expected depth values $d_e$ from the known marker coordinates. The measurements were collected
from the depth range $0.53$\,--\,$0.92\,\text{m}$ in which the objects appear in the training and test images.
We found the following linear correction models by least squares fitting: $d_c = 1.0247 \cdot d - 5.19$ for Carmine, and $d_c = 1.0266 \cdot d - 26.88$ for Kinect (depth measured in $\text{mm}$). In~\cite{sturm2012benchmark}, only scaling is used for the depth correction. According to Foix et al.~\cite{foix2011lock}, a 3-degree polynomial function suffices to correct depth in the $1$\,--\,$2\,\text{m}$ range. In our case,
a narrower range is used and we found a simple linear polynomial to adequately account for the error:
the correction reduced the mean absolute difference from the expected depth $d_e$ from $12.4\,\text{mm}$ to $2.8\,\text{mm}$ for Carmine and from $7.0\,\text{mm}$ to $3.6\,\text{mm}$ for Kinect.
The estimated correction was applied to all depth images, requiring no further action
from the dataset user.




\subsection{3D Object Models}
\label{sec:obj_models}

\begin{figure}[t!]
	\begin{center}
		\begingroup
		\setlength{\tabcolsep}{0.0pt} 
		\renewcommand{\arraystretch}{0.9} 
		\begin{tabular}{ c }
			\includegraphics[width=1.0\columnwidth]{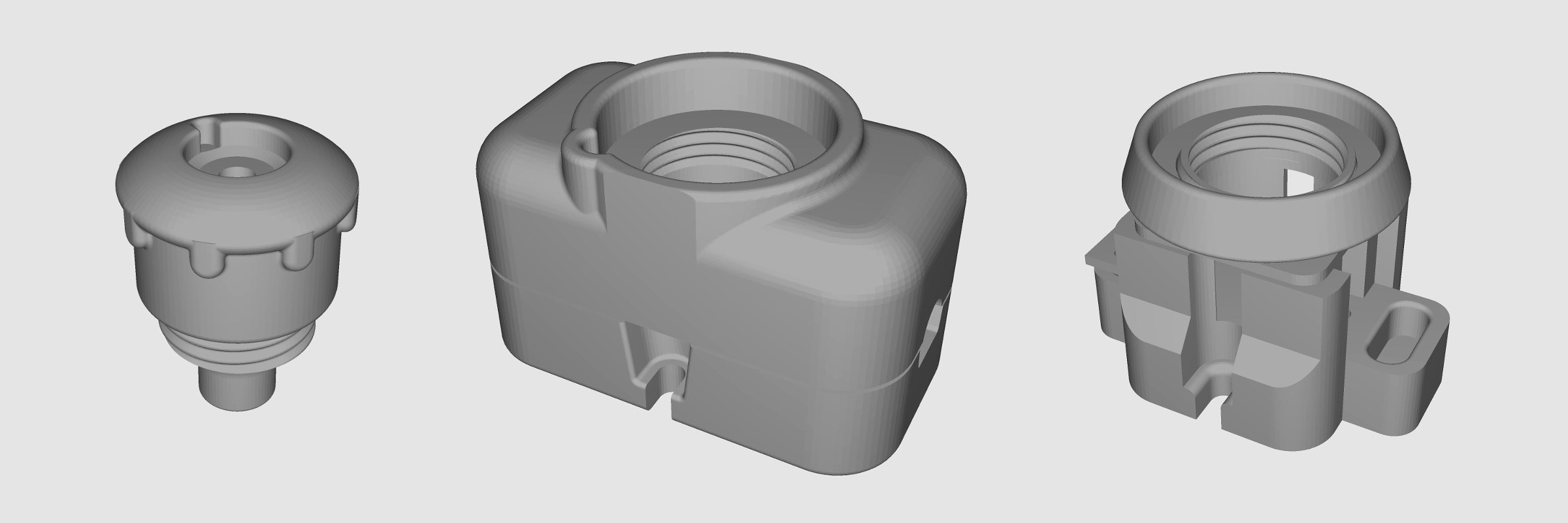} \\
			\includegraphics[width=1.0\columnwidth]{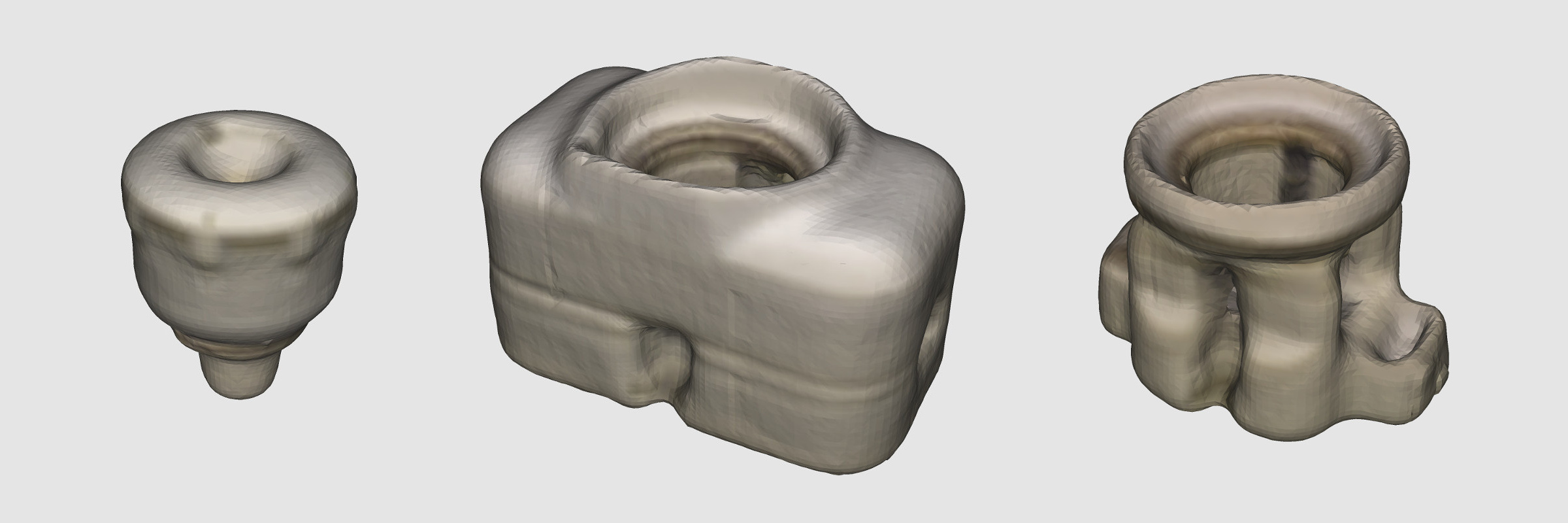}
			
			
		\end{tabular}
		\endgroup
		\caption{\label{fig:obj_models} Examples of 3D object models. Top: Manually created CAD models. Bottom: Semi-automatically reconstructed models which include also surface color. Surface normals at model vertices are included in both model types.}
	\end{center}
\end{figure}

For each object, a manually created CAD model and a semi-automatically reconstructed model are available (Fig.~\ref{fig:obj_models}). Both models are provided in the form of 3D meshes with surface normals at model vertices. Surface color is included only for the reconstructed models. 
The normals were calculated using MeshLab~\cite{cignoni2008meshlab} as the angle-weighted sum of face normals incident to a vertex~\cite{thurrner1998computing}.

The reconstructed models were created using fastfusion, a volumetric 3D mapping system by Steinbr\"ucker et al.~\cite{fastfusion2014}.
The input to fastfusion were the RGB-D training images from Carmine and the associated camera poses estimated using the fiducial markers (see Sec.~\ref{sec:calibration}). For each object, two partial models were first reconstructed, one for the upper and another for the lower view hemisphere. The partial models were then aligned using the iterative closest point (ICP) algorithm applied to their vertices. This was followed by manual refinement that ensured correct registration of surface details that are visible only in color. 
The resulting alignment was applied to the camera poses 
to transform them into a common reference frame, and the updated poses were used to reconstruct the full object model from all images.
These models contained some minor artifacts, \eg~small spikes, which were removed manually. It is noted that some of the objects contain small shiny metal parts whose depth is not reliably captured by the current depth sensors; in general, any glossy or translucent surface is problematic.
Hence, some of these parts, such as the plug poles,
were not reconstructed.

The reconstructed models were aligned to the CAD models using the ICP algorithm and the alignment was further refined manually. Models of both types are therefore defined in the same coordinate system and the provided ground truth poses are valid for both of them.
The origin of the model coordinate system coincides with the center of the bounding box of the CAD model.

The geometrical similarity of the two model types was assessed by calculating the average distance from vertices of the reconstructed models to the closest surface points of the corresponding CAD models. The average distance over all object models was found to be $1.01\,\text{mm}$, which is very low compared to the size of objects that ranges from $58.13\,\text{mm}$ for object 13 to $217.16\,\text{mm}$ for object 8.
Distances in the opposite direction, \ie{} from the CAD models to the reconstructed models, are not informative since some CAD models contain inner parts that are not represented in the reconstructed models. The Metro software by Cignoni et al.~\cite{cignoni1998metro} was used to measure the model differences.

\subsection{Ground Truth Poses}
\label{sec:gt_poses}




To obtain ground truth 6D object poses for images of a test scene, a dense 3D model of the scene was first reconstructed with the system of Steinbr\"ucker et al.~\cite{fastfusion2014}. This was accomplished using all $504$ RGB-D images of the scene along with the sensor poses estimated using the turntable markers.
The CAD object models were then manually aligned to the scene model. To increase accuracy, the object models were rendered into several selected high-resolution scene images from Canon, misalignments were identified and the poses were manually refined accordingly.
This process was repeated until a satisfactory alignment of the renderings with the scene images was achieved.
The final poses were distributed to all test images with the aid of the known camera-to-turntable coordinate transformations.
The transformed poses are provided as the ground truth poses with each test image.



\section{Design Validation and Experiments}
\label{sec:expers}

This section presents an accuracy assessment of the ground truth poses and examines the difficulty of \tless{} with a recent 6D localization method.

\subsection{Accuracy of the Ground Truth Poses}
\label{sec:gtp_accur}

\begin{figure}[!t]
\begin{center}
\begin{tabular}{|l||c|c|c|c|}
  \hline
  Sensor, model type & $\mu_\delta$ & $\sigma_\delta$ & $\mu_{|\delta|}$ & $\mathit{med}_{|\delta|}$ \\
  \hline
  Carmine, CAD & -0.60 & 8.12 & 4.53 & 2.57 \\
  Carmine, reconst. & -0.79 & 7.72 & 4.28 & 2.46 \\
  Kinect, CAD & 4.46 & 11.76 & 8.76 & 5.67 \\
  Kinect, reconst. & 4.08 & 11.36 & 8.40 & 5.45 \\
  \hline
\end{tabular}
\vspace{2.0pt}
\captionof{table}{\label{tab:gt_error} Statistics of differences between the depth of object models at the ground truth poses and the captured depth (in $\text{mm}$). $\mu_\delta$ and $\sigma_\delta$ is the mean and the standard deviation of the differences, $\mu_{|\delta|}$ and $\mathit{med}_{|\delta|}$ is the mean and the median of the absolute differences.}
\vspace{6pt}
\end{center}



\begin{center}
\begingroup
\setlength{\tabcolsep}{0.0pt} 
\renewcommand{\arraystretch}{0.0} 
\begin{tabular}{c}
\includegraphics[width=1.0\columnwidth]{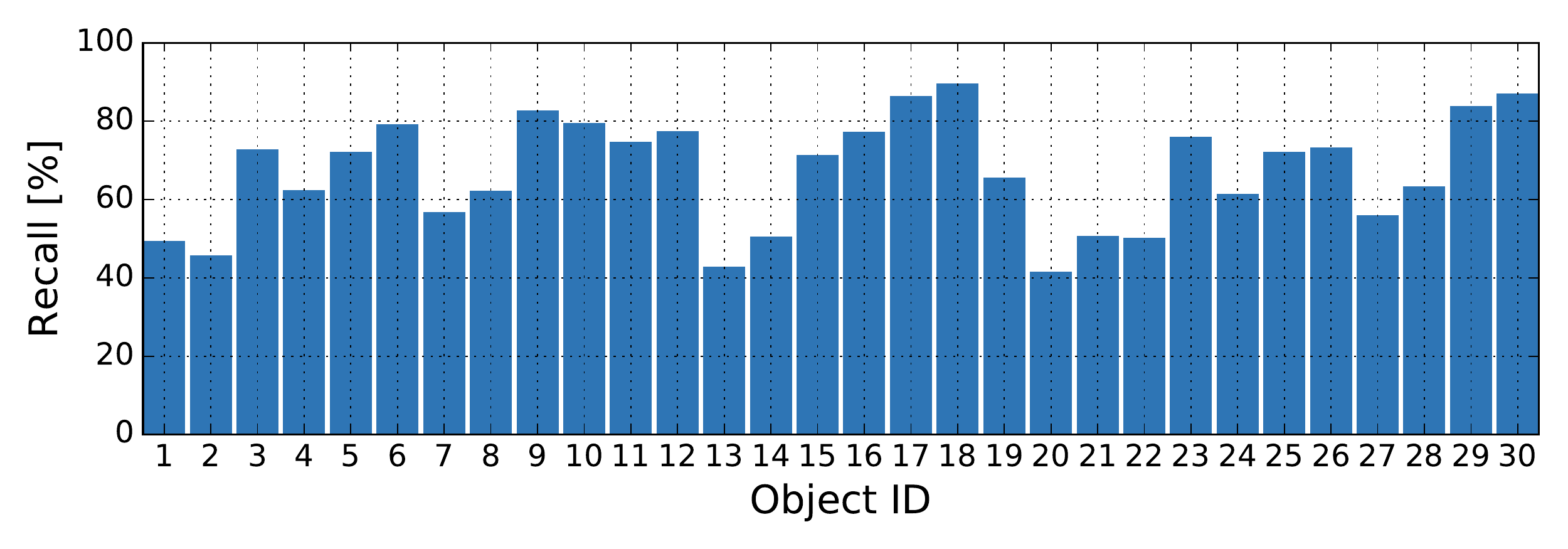} \\
\includegraphics[width=1.0\columnwidth]{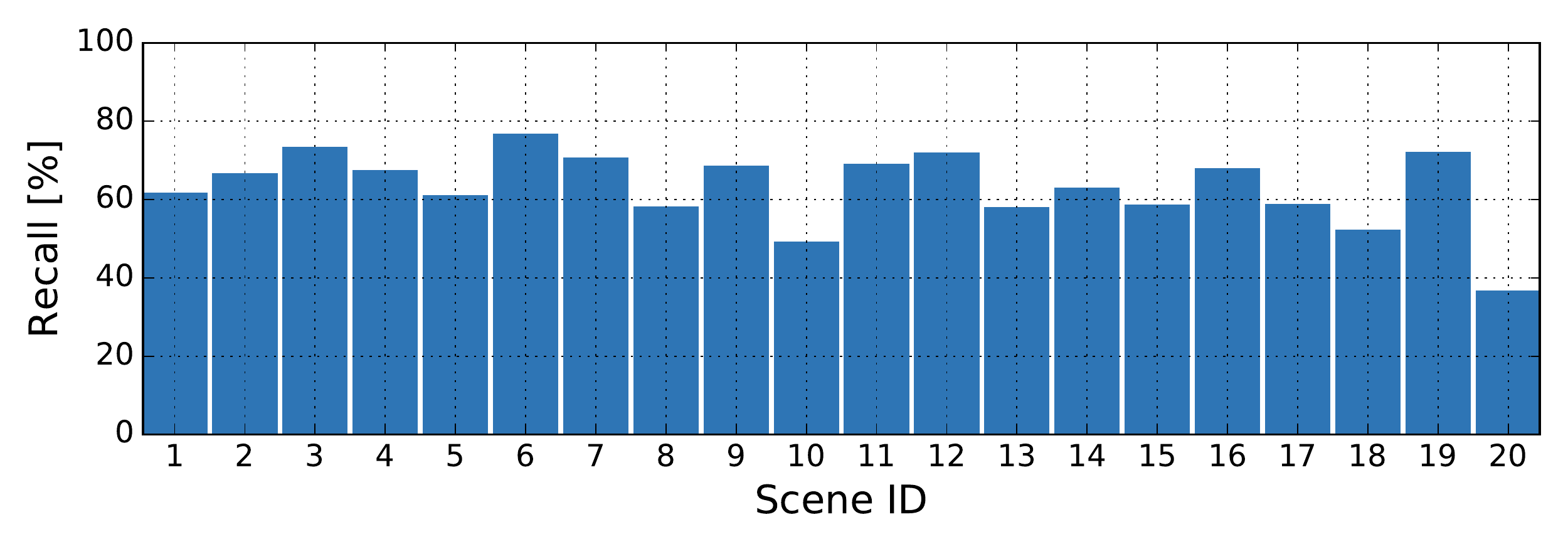} \\ 
\includegraphics[width=1.0\columnwidth]{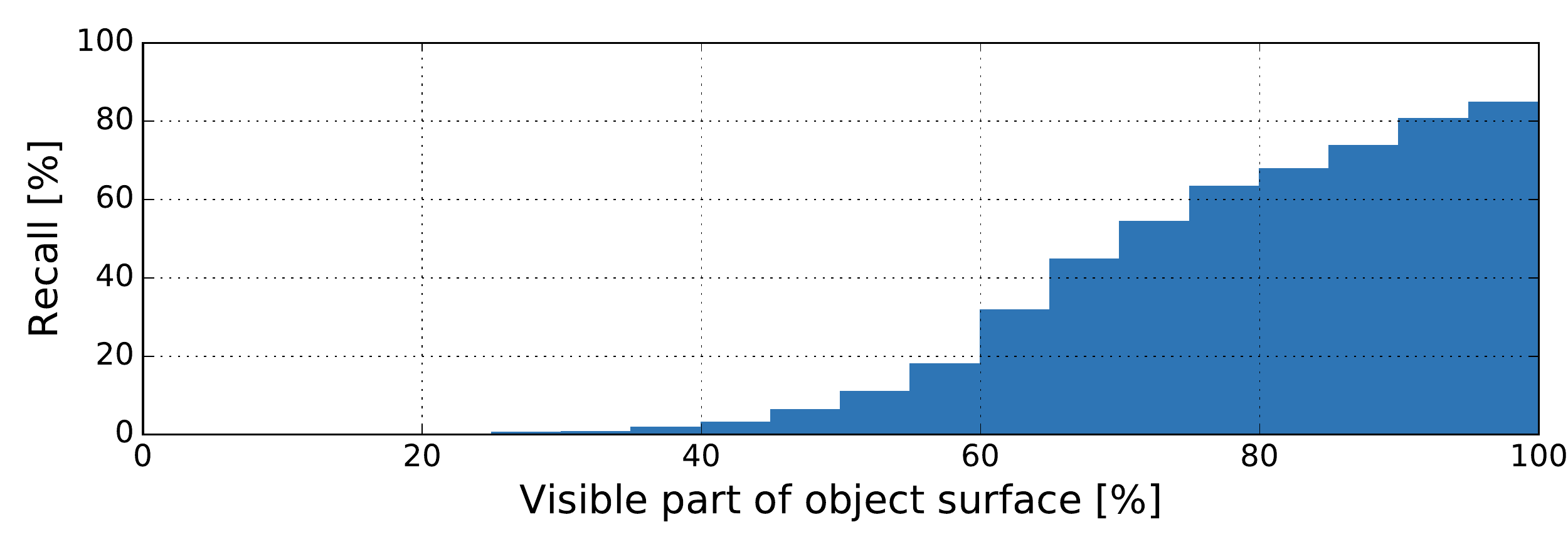}
\end{tabular}
\endgroup
\caption{\label{fig:loc_results} Performance of the method by Hoda\v{n} et al.~\cite{Hodan2015detection} on the 6D localization problem. Shown are the recall per object (top), recall per scene (middle), and recall w.r.t. the percentage of the visible object surface (bottom). 
}
\end{center}
\end{figure}

Aiming to evaluate the accuracy of the ground truth poses, we compared the captured depth images, after the correction described in Sec.~\ref{sec:depth_correction}, with depth images obtained by graphically rendering the 3D object models at the ground truth poses. At each pixel with a valid depth value in both images, we calculated the difference $\delta = d_c - d_r$, where $d_c$ is the captured and $d_r$ is the rendered depth.
Table~\ref{tab:gt_error} presents statistics of these differences,
aggregated over all training and test depth images. Differences exceeding $ 5\,\text{cm}$ and amounting to around $2.5\,\%$ of the measurements were considered to be outliers and were pruned before calculating the statistics. The outlying differences may be caused by erroneous depth measurements, or by occlusion induced by distractor objects in the case of test images.

The rendered depths align well with the depths captured by Carmine, as indicated by the mean difference $\mu_{\delta}$ being close to zero.
In the case of Kinect, we observed that the RGB and depth images are slightly misregistered, which is the cause of the positive bias in $\mu_{\delta}$.
The average absolute difference $\mu_{|\delta|}$ is less than $5\,\text{mm}$ for Carmine and $9\,\text{mm}$ for Kinect, which is near the accuracy of the sensors~\cite{Khoshelham2012} and is relatively small compared to the size of objects. The error statistics are slightly favorable for the
reconstructed models (as opposed to the CAD models), as they were obtained from the captured depth images
and therefore exhibit similar characteristics and artifacts. For example, the plug poles
are invisible to the RGB-D sensors and are missing in the reconstructed models, but are present in the CAD models.


\subsection{6D Localization} \label{sec:6d_loc_eval}

The recent template-based method of Hoda\v{n} et al.~\cite{Hodan2015detection}
was evaluated on the 6D localization problem. The input is comprised of a test image together with the identities of object instances that are present in the image, and the goal is to estimate the 6D poses of these instances~\cite{Hodan2016evaluation}. The method was evaluated on all test RGB-D images from the Carmine sensor. The parameters were set as described in~\cite{Hodan2015detection}, the templates were generated from the training images from Carmine, and the CAD models were employed in the pose refinement stage as detailed in~\cite{Zabulis2015pose}.
Pose estimates were evaluated as in~\cite{hinterstoisser2012model}, using the average distance error
for objects with indistinguishable views.
This error measures the misalignment between the surface of model $\mathcal{M}$ at the ground truth pose $(\bar{\mathbf{R}}, \bar{\mathbf{t}})$ and at the estimated pose $(\hat{\mathbf{R}}, \hat{\mathbf{t}})$, and is defined as:
\begin{equation*}
e = \underset{\mathbf{x}_1 \in \mathcal{M}}{\mathrm{avg}} \; \underset{\mathbf{x}_2 \in \mathcal{M}}{\vphantom{\mathrm{avg}}\mathrm{min}} \left\Vert\big(\bar{\mathbf{R}} \mathbf{x}_1 + \bar{\mathbf{t}}\big) - \big(\hat{\mathbf{R}} \mathbf{x}_2 + \hat{\mathbf{t}}\big)\right\Vert_2.
\end{equation*}


Pose estimate $(\hat{\mathbf{R}}, \hat{\mathbf{t}})$ is considered correct if
$e \leq k \cdot d$,
where $k=0.1$ and $d$ is the largest distance between any pair of model vertices, \ie{} the object diameter. Only the ground truth poses at which at least $10\%$ of the object surface is visible were considered for the evaluation. The visibility was estimated as in~\cite{Hodan2016evaluation}.

The performance is measured by recall, \ie{}~the percentage of the ground truth poses for which a correct pose was estimated.
Fig.~\ref{fig:loc_results} presents achieved recall per object (top) and recall per scene (middle).
The objects with the lowest recall are those that are similar to other objects. For example, object 1 is often confused with object 2, as are objects 20, 21 and 22. Likewise, test scenes containing similar objects are harder, with the hardest one being scene 20 that contains many similar objects and severe occlusions.
Bottom of Fig.~\ref{fig:loc_results} plots the recall, accumulated over all objects, as a function of the fraction of their image projection that is unoccluded.
The recall increases proportionally with this fraction, illustrating that occlusion is one of the main challenges in \tless{}.

The achieved mean recall over all objects is $67.2\%$, which suggests
a significant margin for improvement. We note that the same method achieved a mean recall of $95.4\%$ on the dataset of Hinterstoisser et al.~\cite{hinterstoisser2012model}, which is close to the state of the art: \cite{hinterstoisser2012model} reports $96.6\%$ and \cite{brachmann2016uncertainty} reports $99.0\%$ on this dataset. The latter is not directly comparable since it was calculated only over 13 out of 15 objects included in the dataset.



\section{Conclusion}
\label{sec:concl}

This paper has presented \tless{}, a new dataset for evaluating 6D pose estimation of texture-less objects that can facilitate systematic comparison of pertinent
methods.
The dataset features industry-relevant objects and is characterized by a large number of training and test images, accurate 6D ground truth poses, multiple sensing modalities, test scenes with multiple object instances and with increasing difficulty due to occlusion and clutter.
Initial evaluation results using the dataset indicate that the state of the art in 6D object pose estimation has ample room for improvement. 

\vspace{1em}

\noindent The \tless{} dataset is available online at:\\ \url{cmp.felk.cvut.cz/t-less}



\section*{Acknowledgements}

This work was supported by the Technology Agency of the Czech Republic research program TE01020415 (V3C -- Visual Computing Competence Center),
CTU student grant SGS15/155/OHK3/2T/13,
and the European Commission FP7 DARWIN Project, Grant No. 270138.
The help of Jan Polášek and Avgousta Hatzidaki in creating the CAD models is gratefully acknowledged.

{\small
\bibliographystyle{ieee}
\bibliography{ref}
}

\end{document}